\documentclass[letterpaper, 10 pt, conference]{ieeeconf}  % Comment this line out if you need a4paper

\IEEEoverridecommandlockouts                              % This command is only needed if 
\usepackage{graphics} % for pdf, bitmapped graphics files
\usepackage{epsfig} % for postscript graphics files
\usepackage{mathrsfs}
\usepackage{times} % assumes new font selection scheme installed
\usepackage{amsmath} % assumes amsmath package installed
\usepackage{amssymb}  % assumes amsmath package installed
\usepackage{amsfonts}
\usepackage{amssymb}
\usepackage{pgf}
\usepackage{graphicx}
\usepackage{mathrsfs}
\usepackage{enumerate}
\usepackage{soul}
\usepackage{tikz}
\usepackage{pgfplots}
\usepackage{standalone}
\usepackage{url}
\usepackage{algorithm,algorithmicx}
\usepackage[prependcaption,colorinlistoftodos]{todonotes}

\usepackage{algpseudocode}
\newtheorem{assumption}{Assumption}
\newtheorem{definition}{Definition}

\newcommand{\norm}[1]{\left\lVert#1\right\rVert}

\newcommand{\sN}{\mathcal{N}}

\newcommand{\R}{\mathbb{R}}
\newcommand{\E}{\mathbb{E}}

\newcommand{\bv}[1]{\mathbf{#1}}

\definecolor{gnblue6}{RGB}{35,156,255} % 239cff

\DeclareFontEncoding{LS1}{}{}
\DeclareFontSubstitution{LS1}{stix}{m}{n}
\DeclareSymbolFont{symbols2}{LS1}{stixfrak}{m}{n}
\DeclareSymbolFont{symbols3}{LS1}{stixbb}{m}{n}
\SetSymbolFont{symbols2}{bold}{LS1}{stixfrak}{b}{n}
\SetSymbolFont{symbols3}{bold}{LS1}{stixbb}{b}{n}
\DeclareSymbolFont{arrows1}{LS1}{stixsf}{m}{n}
\SetSymbolFont{arrows1}{bold}{LS1}{stixsf}{b}{n}
\DeclareMathSymbol{\varrightarrow}{\mathrel}{arrows1}{"99}
\DeclareMathSymbol{\varleftarrow}{\mathrel}{arrows1}{"7D}
\DeclareMathSymbol{\fourvdots}{\mathord}{symbols2}{"38}

\newtheorem{thm}{Theorem}[section]

\newtheorem{Problem}{Problem}
\newtheorem{Remark}{Remark}
\title{\LARGE \bf
DR-PETS: Learning-Based Control With Planning in Adversarial Environments
}

\author{Hozefa Jesawada$^{a,1}$, Antonio Acernese$^{b}$, Giovanni Russo$^{a}$, Carmen~Del~Vecchio$^{b}$% <-this % stops a space
\thanks{$^{a}$ Department of Information and Electrical Engineering and Applied Mathematics, University of Salerno, Italy.}
\thanks{$^{b}$  Department of Engineering, University of Sannio, Italy. Corresponding author Carmen Del Vecchio {\tt\small c.delvecchio@unisannio.it}}%
\thanks{$^{1}$ Work done while at the University of Sannio} 
\thanks{The work of HJ and CDV is supported by  PRIN PNRR program of the Italian government CUP C53D23008320001}% <-this % stops a space
}

\begin{document}

\maketitle
\thispagestyle{empty}
\pagestyle{empty}

%%%%%%%%%%%%%%%%%%%%%%%%%%%%%%%%%%%%%%%%%%%%%%%%%%%%%%%%%%%%%%%%%%%%%%%%%%%%%%%%
\begin{abstract}

%Ensuring robustness against epistemic, possibly adversarial, perturbations is critical for reliable planning and decision-making in real-world systems. While the Probabilistic Ensembles with Trajectory Sampling (PETS) algorithm inherently handles certain types of uncertainty through its ensemble-based probabilistic model, its robustness is implicit and limited to "average-case" scenarios, lacking guarantees against structured adversarial perturbations or worst-case uncertainty distributions. To bridge this gap, we propose DR-PETS, a distributionally robust extension of PETS that explicitly equips the algorithm with certified robustness against adversarial uncertainty. By redefining the uncertainty characterization using an ambiguity set based on the $p$-Wasserstein distance, we mathematically formalize adversarial perturbations and enable worst-case-aware planning. While the original PETS framework implicitly accounts for stochasticity, our approach explicitly optimizes for robustness by solving a min-max problem over this Wasserstein ambiguity set. To make the problem tractable, we derive a convex approximation and integrate it into the PETS planning loop. Experiments in control tasks like pendulum stabilization and cart-pole balancing demonstrate that DR-PETS maintains performance under adversarial parameter perturbations, outperforming vanilla PETS in worst-case scenarios while retaining its computational efficiency. This work shows how explicit distributional robustness can augment inherently robust algorithms like PETS for safety-critical applications.

Ensuring robustness against epistemic, possibly adversarial, perturbations is essential for reliable real-world decision-making. While the Probabilistic Ensembles with Trajectory Sampling (PETS) algorithm inherently handles uncertainty via ensemble-based probabilistic models, it lacks guarantees against structured adversarial or worst-case uncertainty distributions. To address this, we propose DR-PETS, a distributionally robust extension of PETS that  certifies robustness against adversarial perturbations. We formalize uncertainty via a $p$-Wasserstein ambiguity set, enabling worst-case-aware planning through a min-max optimization framework. While PETS passively accounts for stochasticity, DR-PETS actively optimizes robustness via a tractable convex approximation integrated into PETS’ planning loop. Experiments on pendulum stabilization and cart-pole balancing show that DR-PETS certifies robustness against adversarial parameter perturbations, achieving consistent performance in worst-case scenarios where PETS deteriorates. %Experiments on pendulum stabilization and cart-pole balancing demonstrate DR-PETS’ superiority under adversarial parameter shifts, outperforming PETS in worst-case settings without sacrificing efficiency. 

%Ensuring robustness against epistemic, possibly adversarial,  perturbations is essential for effective planning and decision-making in real-world environments. In this work, we extend the popular Probabilistic Ensembles with Trajectory Sampling (PETS) algorithm to provide robustness guarantees against adversarial uncertainty distributions, and introduce our distributionally robust DR-PETS algorithm. We characterize the uncertainty using an ambiguity set defined by the $p$-Wasserstein distance. To address the resulting otherwise intractable problem, we propose a tractable approximation and leverage the PETS algorithm to solve the problem. We demonstrate the robust performance of our approach in environments such as the pendulum and cart-pole against parameter perturbations.
%\grtodo{Non ho toccato l'abstract in maniera sostanziale ma lo imposterei in maniera leggermente differente, motivando perche' stiamo estendendo PETS. Ha proprieta' di robustezza intrinseche e quindi lo stiamo equipaggiando in maniera esplicita questa proprieta'?}
\end{abstract}

%%%%%%%%%%%%%%%%%%%%%%%%%%%%%%%%%%%%%%%%%%%%%%%%%%%%%%%%%%%%%%%%%%%%%%%%%%%%%%%%
%%%%%%%%%%%%%%%%%%%%%%%%%%%%%%%%%%%%%%%%%%%%%%%%%%%%%%%%%%%%%%%%%%%%%%%%%%%%%%%%
%%%%%%%%%%%%%%%%%%%%%%%%%%%%%%%%%%%%%%%%%%%%%%%%%%%%%%%%%%%%%%%%%%%%%%%%%%%%%%%%
% PROPOSED INTRO
%%%%%%%%%%%%%%%%%%%%%%%%%%%%%%%%%%%%%%%%%%%%%%%%%%%%%%%%%%%%%%%%%%%%%%%%%%%%%%%%
%%%%%%%%%%%%%%%%%%%%%%%%%%%%%%%%%%%%%%%%%%%%%%%%%%%%%%%%%%%%%%%%%%%%%%%%%%%%%%%%
%%%%%%%%%%%%%%%%%%%%%%%%%%%%%%%%%%%%%%%%%%%%%%%%%%%%%%%%%%%%%%%%%%%%%%%%%%%%%%%%

\section{INTRODUCTION}
Deep reinforcement learning (RL) provides a powerful framework for sequential decision-making. However, deploying RL in practical settings demands robustness to epistemic uncertainty (e.g., adversarial perturbations or model mis-match)~\cite{GARRABE202281}. These uncertainties are often heterogeneous, non-stationary, or even adversarial, making robustness guarantees critical for safe deployment. While recent work has focused on robust RL methods that account for worst-case transitions~\cite{Yu2015}, many state-of-the-art algorithms remain vulnerable to structured adversarial perturbations.

Model-based RL (MBRL) improves sample efficiency by learning an explicit environment model~\cite{pmlr-v168-lellis22a}, enabling planning strategies like model predictive Control (MPC)~\cite{Klink2021} to simulate and optimize long-horizon decisions. Among MBRL methods, the probabilistic ensembles with trajectory sampling (PETS) algorithm~\cite{chua2018} stands out: it combines probabilistic ensemble models with MPC to achieve high data efficiency and performance. However, PETS’ robustness stems implicitly from its ensemble-based uncertainty quantification, which assumes stochastic (not adversarial) deviations. This leaves it ill-equipped for scenarios where perturbations strategically exploit model inaccuracies~\cite{pmlr-v100-okada20a}.

Motivated by this, we propose DR-PETS, a distributionally robust version of PETS. In DR-PETS, planning is still achieved through MPC but, differently from PETS, the MPC cost is regularized and the regularizer arises from a suitable reformulation of the underlying  control problem with ambiguity sets defined by the $p$-Wasserstein distance to capture epistemic uncertainties. Since this leads to an otherwise intractable problem, DR-PETS employs a tractable approximation, resulting in a regularized version of the original  MPC cost. In what follows we briefly survey some closely related works on DR learning and MPC.

\subsubsection*{Related work}
A common approach to handle uncertainty in RL is to consider worst-case transition models~\cite{Iyengar2005}, often formalized via distributionally robust Markov decision processes (DR-MDPs)~\cite{Derman2023,Yu2015}, where ambiguity sets are modeled using the Wasserstein distance~\cite{Gao2023}. Building on this,~\cite{derman2020distributionalregularization} establishes a duality between robustness and regularization, a concept also leveraged in model-free DR RL, such as DR Q-learning~\cite{liu2022DRQ}, which recasts Q-values into a DR Bellman operator via strong duality. Convergence guarantees for Wasserstein-robust Q-learning in linear, time-invariant systems are provided in~\cite{zhao2023minimax}. In~\cite{queeney2023optimal}, a robust RL framework for constrained MDP  using optimal transport perturbations is proposed; however, unlike our method, it requires offline perturbation computation and application to training data. In DR-MPC,~\cite{mark2020stochastic} formulates DR chance constraints and proposes a data-driven approach for value-at-risk under Wasserstein ambiguity, while~\cite{coppens2021datadriven} derives DR-MPC for conic moment-based ambiguity sets, assuming set shrinkage with more data. Finally,~\cite{wu2022ambiguity} proposes an ambiguity tube MPC for nonlinear stochastic systems with cost-to-go concavity.

\subsubsection*{Contributions}
We propose DR-PETS, a DR extension of PETS that explicitly certifies robustness against adversarial perturbations. %To our knowledge, DR-PETS is the first model-based RL method to certify robustness using Wasserstein ambiguity sets while preserving PETS' efficiency. 
Our key contributions include:  
(i) integrating Wasserstein ambiguity sets into the MPC cost functional for worst-case-aware planning without perturbing training data~\cite{derman2020distributionalregularization, queeney2023optimal};  
(ii) reformulating the intractable max-min problem via Wasserstein duality~\cite{blanchet2016quantifying}, yielding a convex maximization with a regularized cost;  
(iii) empirical validation on adversarial pendulum and cart-pole environments, demonstrating superior robustness over PETS;  
(iv) open-source implementation at \url{https://tinyurl.com/595c76t6}.  
Our work advances prior robust RL methods in three key ways: (1) unlike \cite{queeney2023optimal}, which computes adversarial perturbations offline by corrupting training data and then deriving a policy, DR-PETS directly optimizes for a robust policy by maximizing a regularized reward function, eliminating the need for explicit perturbation injection; (2) avoids iterative forward simulations, unlike~\cite{hewing2019cautious}, preserving PETS' data efficiency; and (3) explicitly certifies robustness against adversarial perturbations, unlike PETS, which only accounts for stochastic uncertainty. 
While this  work focuses on theoretical certification of robustness against adversarial perturbations in an unconstrained setting, it establishes a foundational framework for future research to rigorously integrate  safety constraints; indeed, as showed in~\cite{queeney2023optimal}, the inclusion of safety constraints significantly amplifies the benefits of robust methods by reducing sensitivity to model perturbations. This is the objective of future researches.
\section{Preliminaries and problem setup}
%In this section we first provide some preliminaries on the probabilistic ensembles with trajectory sampling (PETS) algorithm by \cite{chua2018}, and explore how the MPC objective can be approximated using samples observed along generated trajectories. Next, we formulate the optimization problem in a distributionally robust setting.
% Finally, some preliminaries on the Wasserstein metric are shown.
% \grtodo{This is largely a copy ad paste of the notation sections we have in the probabilistic papers (except the last few lines). Is it sure that the notation is covered? Already above I found a case where there is undefined notation (see previous comment). We do not deal with pmfs!}
Sets are in {\em calligraphic} and vectors in {\bf bold}. A random variable is denoted by $\bv{V}$ and its realization is $\bv{v}$. We denote the \textit{probability density function}, or simply \textit{pdf}, of $\bv{v}$  by $p(\bv{v})$. Whenever we take the integral involving a pdf we always assume that the integral exists. The  expectation of a function $\mathbf{h}(\cdot)$ of $\bv{v}$ is $\E_{{p}}[\mathbf{h}(\bv{v})]:=\int\mathbf{h}(\bv{v})p(\bv{v})d\bv{\bv{v}}$, where the integration is over the support of $p(\bv{v})$. The {conditional} pdf of $\bv{v}_1$ with respect to $\bv{v}_2$ is $p\left( \bv{v}_1\mid   \bv{v}_2 \right)$. Countable sets are denoted by $w_{k_1:k_n}$, where $w_k$ is the generic set element, $k_1$ ($k_n$) is the index of the first (last) element and  $k_1:k_n$ is the set of consecutive integers between (including) $k_1$ and $k_n$. 
% A pmf of the form $p(\bv{v}_0,\ldots,\bv{v}_N)$ is compactly written as $p_{0:N}$ (by definition $p_{k:k} := p_k(\bv{v}_k)$). We use the shorthand notation $p_{k\mid k-1}$ to denote $p_k(\bv{v}_k\mid\bf{v}_{k-1})$. 
We use $<\cdot,\cdot>$ as the inner product and $\bigotimes$ as the Cartesian product. Let $\mathcal{M}(\mathcal{B})$ represent set of distributions over a Borel set $\mathcal{B}$. Whenever it is clear from the context, a function $g(\cdot)$ is denoted by $g$. The norm is denoted by $\norm{\cdot}$ and the dual norm by $\norm{\cdot}_{*}$.  The symbol $\approx$ denotes approximations.

\subsection{Probabilistic ensembles with trajectory sampling (PETS) algorithm}
Consider a MDP~\cite{chua2018}  defined by the tuple $(\mathcal{X}, \mathcal{U}, f, r, \gamma)$, where $\mathcal{X}$ is the set of states and $\mathcal{U}$ is the set of actions, such that $f: \mathcal{X} \times \mathcal{U} \varrightarrow P(\mathcal{X}) $ is the transition probability function where $P(\mathcal{X})$ denotes space of probability measures over $\mathcal{X}$, $r : \mathcal{X} \times \mathcal{U}  \varrightarrow \mathbb{R}$ is the reward function, and $\gamma \in [0,1)$ is the discount factor. At time $t$, given state $\bv{x}_{t} \in \mathcal{X}\subseteq\R^{x}$, the agent applies action $\bv{u}_{t} \in \mathcal{U}\subseteq\R^{u}$, then the subsequent state $\bv{x}_{t+1}$ is determined by the conditional distribution $f(\bv{x}_{t+1} | \bv{x}_{t}, \bv{u}_{t})$ i.e. $\bv{x}_{t+1}\sim f(\bv{x}_{t+1} | \bv{x}_{t}, \bv{u}_{t}),\forall \bv{x}_{t+1} \in \mathcal{X}$, and $r(\bv{x}_{t},\bv{u}_{t})$  returns the immediate reward received when the MDP is in state $\bv{x}_{t}$ and $\bv{u}_{t}$ is applied. The results in \cite{chua2018}  rely on the following:

\begin{assumption}\label{assm:MDP}
     Given an MDP, $(\mathcal{X}, \mathcal{U}, f, r, \gamma)$ the transition probability function $f$ is unknown and reward $r$ is known.
\end{assumption}

Under Assumption \ref{assm:MDP}, to estimate the transition probability function \( f \), PETS collects a dataset $\mathcal{D}$ of $N \in \mathbb{Z}_{+}$ samples from the system, defined as \(\{(\bv{x}_t, \bv{u}_t, \bv{x}_{t+1})\}_{t=0,1,\dots,N-1}\), where the actions are generated according to some criteria, e. g. randomly. Then an ensemble of \( B \in \mathbb{Z}_{+} \) neural networks (NNs) can be trained on the collected dataset; formally,  the $b$-th NN  is modelled as a Gaussian distribution by $\hat{f}_{\theta_b} \sim \mathcal{N}(m_{\theta_b}, \Sigma_{\theta_b})$, with $b=1,2,\dots,B$ and $\theta_{b}$ being the parameters of $b$-th NN. As in \cite{chua2018}, given a system state $\bv{x}_t$ and an action $\bv{u}_t$, the output of the ensemble, i.e. the probability to reach the new expected state $\hat{\bv{x}}_{t+1}$, is the average over the NNs, i.e., $\hat{f}_{\theta}(\hat{\bv{x}}_{t+1}|\bv{x}_t,\bv{u}_t) := \frac{1}{B} \sum\limits_{b=1}^B \hat{f}_{\theta_b}(\hat{\bv{x}}_{t+1}|\bv{x}_t,\bv{u}_t)$.

For planning and action selection, PETS uses an MPC-based policy. For a given current state $\hat{\bv{x}}_t$, system model $\hat{f}_{\theta}$ and a finite horizon $T\in\mathbb{Z}_{+}$, the RL planning through MPC maximizes the objective function 
\begin{equation}\label{eqn:Chua_obj}
\begin{split}
     \mathcal{J}_{\hat{f}_{\theta}}(\hat{\bv{x}}_t, \bv{u}_{t:t+T-1})=\mathbb{E}_{ \hat{f}_{\theta}}\left[\sum\limits_{i=t}^{t+T-1}\gamma^{i-t}{r}(\hat{\bv{x}}_i,\bv{u}_i) | \hat{\bv{x}}_t\right],
 \end{split}   
\end{equation}
where we denote a simulated state by $\hat{\bv{x}}_{i}$, and for $i=t$ we initialize $\hat{\bv{x}}_t$ to $\bv{x}_t$ as system state. In \eqref{eqn:Chua_obj} the expectation is over simulated future trajectories generated by $\hat{\bv{x}}_{i+1}\sim\hat{f}_{\theta}(\cdot|\hat{\bv{x}}_{i},\bv{u}_{i})$ given the initial state $\bv{x}_{t}$.

With this setting, PETS algorithm aims to find a finite sequence $\bv{u}_{t:t+T-1}^{*} := \{\bv{u}_{t}^{*}, \bv{u}_{t+1}^{*}, \dots, \bv{u}_{t+T-1}^{*}\}$ such that
\begin{equation}\label{eqn:MPC_prob}
\begin{split}
     &\bv{u}_{t:t+T-1}^{\ast} 
     \in \arg \max\limits_{\mathbf{u}_{t:t+T-1}}  \mathcal{J}_{\hat{f}_{\theta}}(\hat{\bv{x}}_t, \bv{u}_{t:t+T-1})\\
     &s.t. \quad \bv{u}_{t:t+T-1}\in\mathcal{U}.
\end{split}
\end{equation}

\subsection{Distributionally robust MDPs}

We introduce distributionally robust MDPs (DR-MDPs) following \cite{derman2020distributionalregularization}, where the unknown transition probability function $f$ is within an uncertainty set $\mathcal{F}$, i.e. $f\in\mathcal{F}$.  As commonly assumed in the literature \cite{queeney2023optimal,derman2020distributionalregularization}, we let $\mathcal{F}$ to  be $(\bv{x},\bv{u})$-rectangular i.e. $\mathcal{F}=\bigotimes_{(\bv{x},\bv{u})\in\mathcal{X}\times\mathcal{U}}\mathcal{F}_{\bv{x},\bv{u}}$ where $\mathcal{F}_{\bv{x},\bv{u}}$ is a set of transitions $f(\cdot|\bv{x},\bv{u})\in P(\mathcal{X})$ for a given state-action pair. Hence, $\mathcal{F}$ is a Borel set and we let any possible transition probability function $\tilde{f}$ be a random variable with support on $\mathcal{F}$, such that $\tilde{f}\sim\tilde{\mu}\in\mathcal{M}(\mathcal{F})$. Here, the class of the distributions $\mathcal{M}$ is the ambuiguity set each of which have support on $\mathcal{F}$.

\begin{definition}\label{def:Wasserstein}($p$-Wasserstein distance \cite{blanchet2016quantifying})
     Let $p\in[1,+\infty]$ and $(\mathcal{F},d)$ be a metric space, where  $d:\mathcal{F}\times\mathcal{F}\rightarrow\mathbb{R}_{+}$ is a lower semicontinuous metric. The $p$-Wasserstein distance  between $\hat{\mu},\tilde{\mu}\in\mathcal{M}$($\mathcal{F}$) is
\begin{equation}\label{eqn:wasserstein}
        \mathcal{W}_{p}(\tilde{\mu},\hat{\mu}) = \inf\limits_{\psi\in\mathcal{Q}(\hat{\mu},\tilde{\mu})}\left(\int_{\mathcal{F}\times\mathcal{F}} d(\hat{f},\tilde{f})^{p} d\psi(\hat{f},\tilde{f})\right)^{\frac{1}{p}},
    \end{equation}
    where $\mathcal{Q}(\hat{\mu},\tilde{\mu})$ is the set of distributions over $\mathcal{F}\times\mathcal{F}$ with marginals $\hat{\mu}$ and $\tilde{\mu}$.
\end{definition}

Inspired by~\cite{derman2020distributionalregularization} and \cite{shi2022distributionally}, we choose a $p$-Wasserstein ball $\Omega_{\epsilon}(\hat{\mu})$ as the ambiguity set over $\mathcal{F}$, centered in $\hat{\mu}$ and defined as 
\begin{equation}\label{eqn:omega}
\Omega_{\epsilon}(\hat{\mu}) := \{\tilde{\mu}: \mathcal{W}_{p}(\tilde{\mu},\hat{\mu}) \leq \epsilon\}
\end{equation}
where $\epsilon>0$ is the radius of the ball, and $\hat{\mu} := \frac{1}{B} \sum\limits_{b=1}^B \delta_{\hat{f}_{\theta_b}}$, with $\delta_{\hat{f}_{\theta_b}}$ being the Dirac distribution with full mass on $\hat{f}_{\theta_{b}}$.

\subsection{Problem statement}
Following \cite{derman2020distributionalregularization}, we let the $p$-Wasserstein DR-MDP (WR-MDP) tuple as $(\mathcal{X},\mathcal{U}, r, \mathcal{F}, \Omega_{\epsilon}(\hat{\mu}))$, and we define corresponding DR formulation of \eqref{eqn:MPC_prob} as: 
\begin{Problem}\label{prob1}

Given a WR-MDP $(\mathcal{X},\mathcal{U}, r, \mathcal{F}, \Omega_{\epsilon}(\hat{\mu}))$, find the sequence of control inputs $\mathbf{u}_{t:t+T-1}^{*} := \{\bv{u}_{t}^{*}, \bv{u}_{t+1}^{*}, \dots, \bv{u}_{t+T-1}^{*}\}$ such that,
\begin{align}\label{eqn:DR_obj}
    &\mathbf{u}_{t:t+T-1}^{*}\in \arg \max\limits_{\mathbf{u}_{t:t+T-1}} \inf_{\tilde{\mu}\in \Omega_{\epsilon}(\hat{\mu})}\mathbb{E}_{\tilde{f}\sim\tilde{\mu}}\left[\mathcal{J}_{\tilde{f}}(\bv{x}_t, \bv{u}_{t:t+T-1})\right] \notag\\
    & s.t. \quad {\bv{u}_t, \bv{u}_{t+1}, \dots, \bv{u}_{t+T-1}} \in \mathcal{U},
\end{align}
    where $\mathcal{J}_{\tilde{f}}(\bv{x}_t, \bv{u}_{t:t+T-1})$ is defined as in \eqref{eqn:Chua_obj}.
    % \[\mathcal{J}_{\tilde{f}}(\bv{x}_t, \bv{u}_{t:t+T-1})=\mathbb{E}_{\tilde{f}}\left[\sum \limits_{i=t}^{t+T-1}\gamma^{i-t}r(\hat{\bv{x}}_i,\bv{u}_i)\mid \bv{x}_{t}\right].\]
\end{Problem}
% \textcolor{blue}{I do not understand the next sentence. What is the problem we are trying to solve? Are we giving a reformulation of (4)?}
By evaluating the infimum over the ambiguity set $\Omega_{\epsilon}(\hat{\mu})$ in \eqref{eqn:DR_obj} we consider a worst-case scenario for the objective function in \eqref{eqn:Chua_obj}, and this corresponds to a robust formulation of \eqref{eqn:MPC_prob}. 
% Thus, we aim to find a lower bound of~\eqref{eqn:Chua_obj}, namely~\eqref{eqn:DR_obj}, that holds with some probability depending on the radius of a Wasserstein ball centered in $\hat{\mu}$.
To lighten the notation, for any given transition probability function $\tilde{f}$ and time sequence $t+i:t+T-1$ starting from current time $t+i$, we denote $\mathcal{J}_{\tilde{f}}(\bv{x}_{t+i}, \bv{u}_{t+i:t+T-1})$ with $\mathcal{J}_{t+i}(\tilde{f})$. In the subsequent analysis we make the following  assumption:
\begin{assumption}\label{assm:linearity} 
Consider Problem \ref{prob1} with $\Omega_{\epsilon}(\hat{\mu})$ defined in~\eqref{eqn:omega}.  The radius of the ball, $\epsilon$, is such that
$
\mathcal{J}_{t}(\bar{f}) \approx \mathcal{J}_{t}(\hat{f}_{\theta_{b}})+\left<\nabla_{\hat{f}_{\theta_{b}}}\mathcal{J}_{t}(\hat{f}_{\theta_{b}}),v_{b}\right>,
$
for all $\bar{f} \in \Omega_{\epsilon}(\hat{\mu})$.
%\AAC{Given the $p$-Wasserstein distance and an adversarial perturbation $v_{b}$  such that $d(\tilde{f}_{b},\hat{f}_{\theta_b})=\norm{v_{b}}$ with $\tilde{f}_{b}=\hat{f}_{\theta_b}+v_{b}$ and $\frac{1}{B}\sum\limits_{b=1}^{B}\norm{v_{b}}^{p}\leq\epsilon^{p}$, 
%then  when $\epsilon$ is sufficiently small.}
\end{assumption}
\begin{Remark}
     Assumption \ref{assm:linearity} is standard in the literature on DR-RL (see e.g.~\cite{derman2020distributionalregularization},~\cite{queeney2023optimal} and~\cite{goodfellow2014explaining}). Intuitively, it implies that $\epsilon$ is sufficiently small, i.e.,  within the feasibility domain of Problem \ref{prob1} a first order approximation for the cost holds.
\end{Remark}
%\grtodo{Ass. 2 and the remark are the major things I changed.}

Authors in \cite{shi2022distributionally} and \cite{queeney2023optimal} showed that the objective function in Problem \ref{prob1} is intractable. Our main objective is to derive a tractable reformulation of cost function in \eqref{eqn:DR_obj}. We do this by leveraging the strong duality of \eqref{eqn:DR_obj}, as established in \cite{blanchet2016quantifying},
\begin{align}\label{eqn:DR_duality}
    &\inf_{\tilde{\mu}\in \Omega_{\epsilon}(\hat{\mu})}\mathbb{E}_{\tilde{f}\sim\tilde{\mu}}\mathcal{J}_{t}(\tilde{f})\notag\\ 
    &= \inf_{\tilde{\mu}} \sup\limits_{\lambda\geq0}\Biggl\{\mathbb{E}_{\tilde{f}\sim\tilde{\mu}}\left[\mathcal{J}_{t}(\tilde{f})\right]\\
    &\qquad \qquad \qquad-\lambda\Biggl(\epsilon^{p}-\frac{1}{B}\sum\limits_{b=1}^{B}\mathbb{E}_{{f_{b}}\sim{\mu_{b}}}\left[d(f_{b},\hat{f}_{\theta_{b}})^{p}\right]\Biggr)\Biggr\}\notag\\
     &= \sup\limits_{\lambda\geq0}\Biggl\{\frac{1}{B}\sum\limits_{b=1}^B\inf_{\tilde{f}_{b}}\left\{\mathcal{J}_{t}(\tilde{f}_{b})+\lambda(d(\tilde{f}_{b},\hat{f}_{\theta_{b}})^{p})\right\}-\lambda\epsilon^{p}\Biggr\}\notag.
\end{align}
Our approach aligns with the DR framework proposed in [11], although  developed for a finite discrete state space.  Indeed,  for our framework of a continuous state space we use of an ensemble of neural networks to model the transition dynamics $f$ and the subsequent derivation of a regularized reward function through a Wasserstein ambiguity set.

\section{Main results}
To develop the DR-PETS algorithm we need a tractable reformulation of \eqref{eqn:DR_obj} for which we employ the last equality in \eqref{eqn:DR_duality}. In fact, in the following Theorem we show that \eqref{eqn:DR_duality} is formed of two terms: the empirical estimate of the performance index over the ensemble and a regularizer. 
% \begin{lem}\label{lem:Jimmy}
% Wasserstein DR policy optimization: derivation using first order approximation.\\
% Consider the $p$-Wasserstein distance, $p \in (1,\infty)$, and let $d(\mathbf{z} + v) = \norm{v}$ for some general norm. From Wasserstein duality, we are interested in developing a tractable reformulation of 
% \[\sup_{\lambda \geq 0}\biggr\{\frac{1}{n}\sum\limits_{i=1}^n[\inf_{\mathbf{z}_i}l(\mathbf{z}_i) + \lambda(d(\mathbf{z}_i, \hat{\mathbf{z}}_i)^p)] - \lambda \epsilon^p\biggl\}.\]
% Using a first order approximation $l(\hat{\mathbf{z}} + v_i) = l(\hat{\mathbf{z}}_i) + \langle g_i, v_i\rangle$, the closed-form
% solution for the Wasserstein DR objective is given by
% \[\frac{1}{n}\sum\limits_{i=1}^n l(\hat{\mathbf{z}}_i) - \epsilon\biggr( \frac{1}{n}\sum\limits_{i=1}^n \norm{g_i}_{\ast}^q\biggl)^{\frac{1}{q}}.\]
% \end{lem}
\begin{thm}\label{thm:Jimmy}
Let Assumption \ref{assm:linearity} hold. Then:
\begin{align}\label{eqn:thm_1_statement}
    &\sup_{\lambda \geq 0}\biggr\{\frac{1}{B}\sum\limits_{b=1}^B\inf_{\tilde{f}_b}
    \bigr[\mathcal{J}_{t}(\tilde{f}_{b}) + \lambda(d(\tilde{f}_b, \hat{f}_{\theta_{b}})^{p})\bigl] - \lambda \epsilon^p\biggl\}\notag
    \\
    &=\frac{1}{B}\sum\limits_{b=1}^B \mathcal{J}_{t}(\hat{f}_{\theta_{b}}) - \epsilon\biggr( \frac{1}{B}\sum\limits_{b=1}^B \norm{\nabla_{\hat{f}_{\theta_{b}}}\mathcal{J}_{t}(\hat{f}_{\theta_{b}})}_{\ast}^{\frac{p}{p-1}}\biggl)^{\frac{p-1}{p}}.
\end{align}

% where let $l(\hat{f}_{\theta_{b}}) = \mathbb{E}_{\hat{f}_{\theta_{b}}}\biggr[\sum \limits_{i=t}^{t+T-1}\gamma^{i-t}r(\hat{\bv{x}}_i,\bv{u}_i)\biggl]$ and $g_{b}(\hat{f}_{\theta_{b}}) = \nabla_{\hat{f}_{\theta_{b}}}\mathbb{E}_{\hat{f}_{\theta_{b}}}\bigr[\sum \limits_{i=t}^{t+T-1}\gamma^{i-t}r(\hat{\bv{x}}_i,\bv{u}_i)\bigl].$
\end{thm}

\begin{proof}
Motivated by the approach of \cite{goodfellow2014explaining} and relying on Assumption \ref{assm:linearity} we can write the last equality in \eqref{eqn:DR_duality} as
 \begin{align}\label{eqn:dual_problem_recast}
             &\sup_{\lambda \geq 0}\biggr\{\frac{1}{B}\sum\limits_{b=1}^B\inf_{\tilde{f}_b}
    \left[\mathcal{J}_{t}(\tilde{f}_{b}) + \lambda(d(\tilde{f}_b, \hat{f}_{\theta_{b}})^{p})\right] - \lambda \epsilon^p\biggl\}\notag\\
    % &=\sup_{\lambda \geq 0}\biggr\{\frac{1}{B}\sum\limits_{b=1}^B\inf_{v_{b}}\left[\mathcal{J}_{t}(\hat{f}_{\theta_{b}}+v_{b}) + \lambda(\norm{v_{b}}^{p})\right] - \lambda \epsilon^p\biggl\}\\
    &\approx\frac{1}{B}\sum\limits_{b=1}^{B}\mathcal{J}_{t}(\hat{f}_{\theta_{b}})+\sup_{\lambda \geq 0}\biggr\{\frac{1}{B}\sum\limits_{b=1}^B\inf_{v_{b}}\Bigl[\left<\nabla_{\hat{f}_{\theta_{b}}}\mathcal{J}_{t}(\hat{f}_{\theta_{b}}),v_{b}\right>\notag\\ 
    & \qquad \qquad + \lambda(\norm{v_{b}}^{p})\Bigr] - \lambda \epsilon^p\biggl\}.
 \end{align}
% \grtodo{Note that I changed to approximation the last line.}
% By plugging \eqref{eqn:first_order_approx} into \eqref{eqn:dual_problem}, we obtain
% \begin{equation}\label{eqn:dual_problem_recast}
%     \begin{split}
%     &\sup_{\lambda \geq 0}\biggr\{\frac{1}{B}\sum\limits_{b=1}^B\inf_{v_{b}}
%     \left[\mathcal{J}_{t}(\hat{f}_{\theta_{b}}+v_{b}) + \lambda(\norm{v_{b}}^{p})\right] - \lambda \epsilon^p\biggl\}\\
%     &=\frac{1}{B}\sum\limits_{b=1}^{B}\mathcal{J}_{t}(\hat{f}_{\theta_{b}})+\sup_{\lambda \geq 0}\biggr\{\frac{1}{B}\sum\limits_{b=1}^B\inf_{v_{b}}\Bigl[\left<\nabla_{\hat{f}_{\theta_{b}}}\mathcal{J}_{t}(\hat{f}_{\theta_{b}}),v_{b}\right>\\ 
%     & \qquad \qquad + \lambda(\norm{v_{b}}^{p})\Bigr] - \lambda \epsilon^p\biggl\}.
%     \end{split}
%  \end{equation}

We begin our proof by solving the inner minimization problem in \eqref{eqn:dual_problem_recast}; we define $j(v_b)$ as
\begin{equation}\label{eqn:inner_prob}
\inf_{v_{b}} j(v_{b}) := \inf\limits_{v_{b}}\left\{\left<\nabla_{\hat{f}_{\theta_{b}}}\mathcal{J}_{t}(\hat{f}_{\theta_{b}}),v_{b}\right>+\lambda\norm{v_{b}}^{p}\right\},
\end{equation}
% Note that value of $\lambda$ determined by the solving the outer maximization in \eqref{eqn:dual_problem_recast} (\AAC{what do you mean?}).
where we introduce the auxiliary variable $\delta_{b}$, such that $\norm{v_{b}}\geq\delta_{b}$, obtaining
\begin{equation}\label{eqn:inner_prob_aux}
\begin{split}
&\inf\limits_{v_{b},\delta_{b}}\left\{\left<\nabla_{\hat{f}_{\theta_{b}}}\mathcal{J}_{t}(\hat{f}_{\theta_{b}}),v_{b}\right>+\lambda\delta_{b}^{p}\mid\norm{v_{b}}\leq\delta_{b}\right\}\\
&=\inf_{v_{b},\delta_{b}}\left\{h(v_{b},\delta_{b})\mid\norm{v_{b}}\leq\delta_{b}\right\}.
\end{split}
\end{equation}
First, note that for any $v_{b}$, $j(v_{b})\leq h(v_{b},\delta_{b})$ for all feasible $\delta_{b}$ in \eqref{eqn:inner_prob_aux}. This implies $j(v_{b})\leq\inf_{\delta_{b}}\left\{h(v_{b},\delta_{b}\mid\norm{v_{b}}\leq\delta_{b}) \right\}$, and by minimizing both sides w.r.t $v_{b}$ we see that \[\inf_{v_{b}} j(v_{b})\leq\inf_{v_{b},\delta_{b}}\left\{h(v_{b},\delta_{b})\mid\norm{v_{b}}\leq\delta_{b}\right\}.\] Let $\delta_{b}^{*}=\norm{v_{b}^{*}}$, where $v_{b}^{*}$ is the optimal solution to \eqref{eqn:inner_prob}, and note that $(v_{b}^{*},\delta_{b}^{*})$ is a feasible solution to \eqref{eqn:inner_prob_aux} such that $h(v_{b}^{*},\delta_{b}^{*})=j(v_{b}^{*})$. Therefore, we can derive the optimal solution of \eqref{eqn:inner_prob} by finding the optimal solution of \eqref{eqn:inner_prob_aux}:
\begin{align}\label{eqn:delta_opt}
    &\inf\limits_{v_{b},\delta_{b}}\left\{\left<\nabla_{\hat{f}_{\theta_{b}}}\mathcal{J}_{t}(\hat{f}_{\theta_{b}}),v_{b}\right>+\lambda\delta_{b}^{p}\mid\norm{v_{b}}\leq\delta_{b}\right\}\notag\\
    &= \inf\limits_{\delta_{b}\geq0}\left\{\inf\limits_{v_{b}}\left\{\left<\nabla_{\hat{f}_{\theta_{b}}}\mathcal{J}_{t}(\hat{f}_{\theta_{b}}),v_{b}\right>+\lambda\delta_{b}^{p}\mid\norm{v_{b}}\leq\delta_{b}\right\}\right\}\notag\\
    &= \inf\limits_{\delta_{b}\geq0}\left\{\lambda\delta_{b}^{p}+\inf\limits_{\norm{v_{b}}\leq\delta_{b}}\left\{\left<\nabla_{\hat{f}_{\theta_{b}}}\mathcal{J}_{t}(\hat{f}_{\theta_{b}}),v_{b}\right>\right\}\right\}\notag\\
    % &= \inf\limits_{\delta_{b}\geq0}\left\{\lambda\delta_{b}^{p}-\sup\limits_{\norm{v_{b}}\leq\delta_{b}}\left\{\left<\nabla_{\hat{f}_{\theta_{b}}}\mathcal{J}_{t}(\hat{f}_{\theta_{b}}),v_{b}\right>\right\}\right\}\\
    &= \inf\limits_{\delta_{b}\geq0}\left\{\lambda\delta_{b}^{p}-\delta_{b}\norm{\nabla_{\hat{f}_{\theta_{b}}}\mathcal{J}_{t}(\hat{f}_{\theta_{b}})}_{*}\right\},
\end{align}
where the last step comes from the definition of the dual norm $\norm{\cdot}_{*}$. The minimization over $\delta_{b}$ in \eqref{eqn:delta_opt} is dependent on $\lambda$ and the value $\delta_{b}^{*}(\lambda)$ that achieves the minimum can be obtained by exploiting the convexity of \eqref{eqn:delta_opt} in $\delta_{b}$. To obtain $\delta_{b}^{*}(\lambda)$
 we set the derivative of $\lambda\delta_{b}^{p}-\delta_{b}\norm{\nabla_{\hat{f}_{\theta_{b}}}\mathcal{J}_{t}(\hat{f}_{\theta_{b}})}_{*}$ with respect to $\delta_{b}$ equal to zero, which leads to
% \begin{equation}\label{eqn:delta_div}
% p\lambda\delta_{b}^{p-1}-\norm{\nabla_{\hat{f}_{\theta_{b}}}\mathcal{J}_{t}(\hat{f}_{\theta_{b}})}_{*}
% \end{equation}
% Solving \eqref{eqn:delta_div} for $\delta_{b}$ results in $\delta^{*}_{b}(\lambda)$ as,
\begin{equation*}\label{eqn:delta_star}
\delta_{b}^{*}(\lambda)=\left(\frac{\norm{\nabla_{\hat{f}_{\theta_{b}}}\mathcal{J}_{t}(\hat{f}_{\theta_{b}})}_{*}}{p\lambda}\right)^{\frac{1}{p-1}}.
\end{equation*}

By plugging $\delta_{b}^{*}(\lambda)$ in \eqref{eqn:delta_opt} we obtain a solution to the inner minimization in~\eqref{eqn:dual_problem_recast}:

\begin{equation}\label{eqn:inner_sol}
    \begin{split}
&\inf\limits_{v_{b}}\left[\left<\nabla_{\hat{f}_{\theta_{b}}}\mathcal{J}_{t}(\hat{f}_{\theta_{b}}),v_{b}\right>+\lambda\norm{v_{b}}^{p}\right]\\ 
&= \lambda(\delta_{b}^{*}(\lambda))^{p}-\delta_{b}^{*}(\lambda)\norm{\nabla_{\hat{f}_{\theta_{b}}}\mathcal{J}_{t}(\hat{f}_{\theta_{b}})}_{*}\\
& = \frac{(1-p)\norm{\nabla_{\hat{f}_{\theta_{b}}}\mathcal{J}_{t}(\hat{f}_{\theta_{b}})}_{*}^{\frac{p}{p-1}}}{p^{\frac{p}{p-1}}\lambda^{\frac{1}{p-1}}}. 
    \end{split}
\end{equation}

We can now solve the outer maximization in \eqref{eqn:dual_problem_recast} by using  \eqref{eqn:inner_sol} which leads to
\begin{align}\label{eqn:outer_prob}
    &\sup_{\lambda \geq 0}\biggr\{\frac{1}{B}\sum\limits_{b=1}^B\inf_{v_{b}}\left[\left<\nabla_{\hat{f}_{\theta_{b}}}\mathcal{J}_{t}(\hat{f}_{\theta_{b}}),v_{b}\right> + \lambda(\norm{v_{b}}^{p})\right] - \lambda \epsilon^p\biggl\}\notag\\
    &=\sup_{\lambda \geq 0}\left\{\frac{(1-p)\left(\frac{1}{B}\sum_{b=1}^{B}\norm{\nabla_{\hat{f}_{\theta_{b}}}\mathcal{J}_{t}(\hat{f}_{\theta_{b}})}_{*}^{\frac{p}{p-1}}\right)}{p^{\frac{p}{p-1}}\lambda^{\frac{1}{p-1}}} - \lambda \epsilon^p\right\}.
\end{align}
Note that \eqref{eqn:outer_prob} is a maximization of a concave function in $\lambda$: the solution can be obtained by setting the derivative of~\eqref{eqn:outer_prob} w.r.t. $\lambda$ equal to zero:
\begin{align}\label{eqn:grad_lambda}
        &\frac{\mathrm{d}}{\mathrm{d}\lambda}\left(\frac{(1-p)\left(\frac{1}{B}\sum_{b=1}^{B}\norm{\nabla_{\hat{f}_{\theta_{b}}}\mathcal{J}_{t}(\hat{f}_{\theta_{b}})}_{*}^{\frac{p}{p-1}}\right)}{p^{\frac{p}{p-1}}\lambda^{\frac{1}{p-1}}} - \lambda \epsilon^p\right)\notag\\
        &=\frac{\left(\frac{1}{B}\sum_{b=1}^{B}\norm{\nabla_{\hat{f}_{\theta_{b}}}\mathcal{J}_{t}(\hat{f}_{\theta_{b}})}_{*}^{\frac{p}{p-1}}\right)}{p^{\frac{p}{p-1}}\lambda^{\frac{p}{p-1}}} -  \epsilon^{p} .
    \end{align}
Setting \eqref{eqn:grad_lambda} equal to zero and solving for $\lambda$ gives,
\begin{equation*}\label{eqn:lambda_star}
    \lambda^{*}=\frac{\left(\frac{1}{B}\sum_{b=1}^{B}\norm{\nabla_{\hat{f}_{\theta_{b}}}\mathcal{J}_{t}(\hat{f}_{\theta_{b}})}_{*}^{\frac{p}{p-1}}\right)^{\frac{p-1}{p}}}{p\epsilon^{p-1}}.
\end{equation*}
By plugging $\lambda^{*}$ into \eqref{eqn:outer_prob} we have
\begin{equation}\label{eqn:outer_sol}
    \begin{split}
        % &\sup_{\lambda \geq 0}\left\{\frac{(1-p)\left(\frac{1}{B}\sum_{b=1}^{B}\norm{\nabla_{\hat{f}_{\theta_{b}}}\mathcal{J}_{t}(\hat{f}_{\theta_{b}})}_{*}^{\frac{p}{p-1}}\right)}{p^{\frac{p}{p-1}}\lambda^{\frac{1}{p-1}}} - \lambda \epsilon^p\right\}\\
        &\frac{(1-p)\left(\frac{1}{B}\sum_{b=1}^{B}\norm{\nabla_{\hat{f}_{\theta_{b}}}\mathcal{J}_{t}(\hat{f}_{\theta_{b}})}_{*}^{\frac{p}{p-1}}\right)}{p^{\frac{p}{p-1}}\lambda^{*\frac{1}{p-1}}} - \lambda^{*} \epsilon^p\\
        &=-\epsilon\left(\sum_{b=1}^{B}\norm{\nabla_{\hat{f}_{\theta_{b}}}\mathcal{J}_{t}(\hat{f}_{\theta_{b}})}_{*}^{\frac{p}{p-1}}\right)^{\frac{p-1}{p}}.
    \end{split}
\end{equation}
Finally, combining \eqref{eqn:outer_sol} with \eqref{eqn:dual_problem_recast} we arrive at \eqref{eqn:thm_1_statement} as the closed form solution.\end{proof}
% Using Jimmy's results given in lemma \ref{lem:Jimmy}, with

% \begin{equation*}
%     \begin{cases}
%         p=2\\
%         \mathbf{z} = \tilde{f}\\
%         \hat{\mathbf{z}}_i = \hat{f}_{\theta_i}\\
%         l(\mathbf{z}) = \mathbb{E}_{\tilde{f}}\biggr[\sum \limits_{i=t}^{t+T-1}\gamma^{i-t}r(\hat{\bv{x}}_i,\bv{u}_i)\biggl]\\
%         g_i = \nabla_{\tilde{f}}\mathbb{E}_{\tilde{f}}\bigr(\sum \limits_{i=t}^{t+T-1}\gamma^{i-t}r(\hat{\bv{x}}_i,\bv{u}_i)\bigl)\biggr|_{\tilde{f}=\hat{f_b}}\\
%         n = B
%     \end{cases},
% \end{equation*}

% and relying on Wasserstein duality, we can derive a regularized version of the RHS of \eqref{eqn:DR_start}, namely:

% \begin{equation}\label{eqn:Jimmy}
%     \mathcal{J}_{\Omega_{\epsilon}(\hat{\mu})}(\bv{x}_t, \bv{u}_{t:t+T-1}) \approx \frac{1}{B}\sum\limits_{b=1}^B \mathbb{E}_{\hat{f}_{\theta_b} }\biggr(\sum \limits_{i=t}^{t+T-1}\gamma^{i-t}r(\hat{\bv{x}}_i,\bv{u}_i)\biggl) -\epsilon \biggr(\frac{1}{B}\sum\limits_{b=1}^B \norm{\nabla_{\tilde{f}}\mathbb{E}_{\tilde{f}}\bigr(\sum \limits_{i=t}^{t+T-1}\gamma^{i-t}r(\hat{\bv{x}}_i,\bv{u}_i)\bigl)\biggr|_{\tilde{f}=\hat{f}_{\theta_b}}}^2\biggl)^{\frac{1}{2}}.
% \end{equation}
As in PETS algorithm, we compute the expectation over the estimated state trajectory in \eqref{eqn:Chua_obj} using a particle-based propagation, i.e., propagating $Q \in \mathbb{Z}_{+}$ particles from initial state $\hat{\bv{x}}^\text{q}_{t}=\bv{x}_{t}$ at time $t$, where $\text{q} = 1,2,\dots,Q$. Each particle is propagated using $\hat{\bv{x}}^{\text{q}}_{t+1}\sim\hat{f}_{\theta_{b}}$. Furthermore, as in PETS, we use cross entropy method (CEM) \cite{botev2013cross} to select actions, which iteratively samples $M \in \mathbb{Z}_{+}$ action sequences $\{\bv{u}_{t:t+T-1}^i\}_{i=1,2,\dots,M}$ from a candidate distribution $\mathbf{CEM}(\cdot):=\sN(\varphi_{\bv{u}},\Sigma_{\bv{u}})$. Actions sampled from the candidate distribution are evaluated using  \eqref{eqn:thm_1_statement}, and parameters $\varphi_{\bv{u}}$ and  $\Sigma_{\bv{u}}$ are updated iteratively based on best action sequence among $M$ samples.

 By exploiting the B-NNs ensemble and letting $p=2$ we develop a simplified version of \eqref{eqn:thm_1_statement} that we use in our DR-PETS algorithm that is reported in the following result.

\begin{thm}\label{thm:main}
    Given the problem in \eqref{eqn:DR_obj}, the optimization objective can be reformulated as,  
    \begin{equation}\label{eqn:thm_statement}
    \begin{split}
       % &\mathcal{J}_{\Omega_{\epsilon}(\hat{\mu})}(\bv{x}_t, \bv{u}_{t:t+T-1}) \approx 
       &\max_{\bv{u}_{t:t+T-1}}\biggr\{ \frac{1}{B}\sum\limits_{b=1}^B \mathbb{E}_{\hat{f}_{\theta_b} }\biggr[\sum \limits_{i=t}^{t+T-1}\gamma^{i-t}r(\hat{\bv{x}}_i,\bv{u}_i)\biggl]\\ & -\epsilon \biggr(\frac{1}{B}\sum\limits_{b=1}^B \bigg\|
    \sum\limits_{k=1}^{T-1} \gamma^{k} \mathbb{E}_{\hat{\bv{x}}_{t+1:t+k}\sim\hat{f}_{\theta_{b}}}\\
    &\biggr[ \nabla_{\hat{f}_{\theta_{b}}} \ln(\hat{f}_{\theta_{b}}(\hat{\bv{x}}_{t+k}|\hat{\bv{x}}_{t+k-1}, \bv{u}_{t+k-1}))\\
    &\sum\limits_{i=t+k}^{t+T-1} \mathbb{E}_{\hat{\bv{x}}_{i+1}\sim\hat{f}_{\theta_{b}}}\bigr[\gamma^{i-t-k} r(\hat{\bv{x}}_{i}, \bv{u}_{i})\bigl] \biggl]\bigg\|^2\biggl)^{\frac{1}{2}}\biggl\}\\
     & s.t. \quad {\bv{u}_t, \bv{u}_{t+1}, \dots, \bv{u}_{t+T-1}} \in \mathcal{U}.
    \end{split}
    \end{equation}
\end{thm}

\begin{proof}
Considering Theorem \ref{thm:Jimmy}, when $p=2$, the l.h.s. of \eqref{eqn:thm_1_statement} can be rewritten as,
% \begin{equation*}
%     \begin{cases}
%         p=2\\
%         \mathbf{z} = \tilde{f}\\
%         \hat{\mathbf{z}}_i = \hat{f}_{\theta_i}\\
%         l(\mathbf{z}) = \mathbb{E}_{\tilde{f}}\biggr[\sum \limits_{i=t}^{t+T-1}\gamma^{i-t}r(\hat{\bv{x}}_i,\bv{u}_i)\biggl]\\
%         g_i = \nabla_{\tilde{f}}\mathbb{E}_{\tilde{f}}\bigr(\sum \limits_{i=t}^{t+T-1}\gamma^{i-t}r(\hat{\bv{x}}_i,\bv{u}_i)\bigl)\biggr|_{\tilde{f}=\hat{f_b}}\\
%         n = B,
%     \end{cases}
% \end{equation*}
\begin{equation}\label{eqn:Jimmy}
\begin{split}
    % \mathcal{J}_{\Omega_{\epsilon}(\hat{\mu})}(\bv{x}_t, \bv{u}_{t:t+T-1}) \approx
    &\frac{1}{B}\sum\limits_{b=1}^B \mathcal{J}_{t}(\hat{f}_{\theta_{b}})-\epsilon \left(\frac{1}{B}\sum\limits_{b=1}^B \norm{\nabla_{\hat{f}_{\theta_{b}}}\mathcal{J}_{t}(\hat{f}_{\theta_{b}})}^2\right)^{\frac{1}{2}}.
\end{split}
\end{equation}

Now, considering the horizon $T$, during the planning we sample $M$ possibly different sequences of actions, i.e., $\{\bv{u}_{t:t+T-1}^k\}_{k=1,2,\dots,M}$. Thus, the expectation in the first part of \eqref{eqn:Jimmy} is only w.r.t. $\hat{f}_{\theta_{b}}$, and it can be approximated by propagating $Q$ particles for each $k$. Next, we simplify the gradient term $\nabla_{\hat{f}_{\theta_{b}}}\mathcal{J}_{t}(\hat{f}_{\theta_{b}})$ in \eqref{eqn:Jimmy} by unrolling it over time horizon $T$. We use $\hat{\bv{x}}$ to emphasize that the state is predicted using the learned model. For brevity consider $\mathcal{J}_{t}=\mathcal{J}_{t}(\hat{f}_{\theta_{b}})$.
\begin{equation}\label{eqn:gradientVF0}
\begin{split}
    &\nabla_{\hat{f}_{\theta_{b}}}\mathcal{J}_{t}
    = \nabla_{\hat{f}_{\theta_{b}}}\biggr[\sum \limits_{i=t}^{t+T-1}\mathbb{E}_{\hat{f}_{\theta_{b}}}\bigr[\gamma^{i-t}r(\hat{\bv{x}}_i,\bv{u}_i)\bigl]\biggl]\\
    % &=\nabla_{\tilde{f}}\biggr[  r({x}_t,\bv{u}_t) + \gamma\int \tilde{f}({x}_{t+1}|{x}_t,\bv{u}_t) \sum \limits_{i=t+1}^{t+T-1}\mathbb{E}_{\tau\sim (\pi,\tilde{f})}\bigr[\gamma^{i-t-1}r({x}_i,\bv{u}_i)\bigl]d{{x}_{t+1}}\biggl],\\
    % &=\nabla_{\tilde{f}}\biggr[  r({x}_t,\bv{u}_t) + \gamma\int \tilde{f}({x}_{t+1}|{x}_t,\bv{u}_t) \mathcal{J}_{\tilde{f}}({x}_{t+1}, \bv{u}_{t+1:t+T-1})d{{x}_{t+1}}\biggl],\\
    &= \gamma\int\nabla_{\hat{f}_{\theta_{b}}}\biggr( \hat{f}_{\theta_{b}}(\hat{\bv{x}}_{t+1}|\hat{\bv{x}}_t,\bv{u}_t) \mathcal{J}_{t+1}\biggl)d{\hat{\bv{x}}_{t+1}}\\
    &= \phi(\hat{\bv{x}}_t,\bv{u}_t) + \gamma\int \hat{f}_{\theta_{b}}(\hat{\bv{x}}_{t+1}|\hat{\bv{x}}_t,\bv{u}_t) \nabla_{\hat{f}_{\theta_{b}}}\left(\mathcal{J}_{t}\right)d{\hat{\bv{x}}_{t+1}},
\end{split}
\end{equation}
where
$%\begin{equation*}
    \phi(\hat{\bv{x}}_t,\bv{u}_t) = \gamma\int \nabla_{\hat{f}_{\theta_{b}}}\bigr( \hat{f}_{\theta_{b}}(\hat{\bv{x}}_{t+1}|\hat{\bv{x}}_t,\bv{u}_t)\bigl) \mathcal{J}_{t}d{\hat{\bv{x}}_{t+1}}.
$%\end{equation*}

Continuing the unrolling of \eqref{eqn:gradientVF0} with $\hat{f}_{\theta_{b}}(\hat{\bv{x}}_{t+i}) = \hat{f}_{\theta_{b}}(\hat{\bv{x}}_{t+i}|\hat{\bv{x}}_{t+i-1}, \bv{u}_{t+i-1})$ we have:

\begin{align}\label{eqn:gradientVF}
    &\nabla_{\hat{f}_{\theta_{b}}}\mathcal{J}_{t}
    % &= \phi({x}_t,\bv{u}_t) + \gamma\int \tilde{f}({x}_{t+1}|{x}_t,\bv{u}_t) \nabla_{\tilde{f}}\mathcal{J}_{\tilde{f}}({x}_{t+1}, \bv{u}_{t+1:t+T-1})d{{x}_{t+1}},\\
    = \phi(\hat{\bv{x}}_t,\bv{u}_t) + \gamma\int \hat{f}_{\theta_{b}}(\hat{\bv{x}}_{t+1}) \nabla_{\hat{f}_{\theta_{b}}}\mathcal{J}_{t+1}d\hat{\bv{x}}_{t+1}\notag\\
    % &= \phi({x}_t,\bv{u}_t) + \gamma\int f({x}_{t+1})\nabla_{f}\biggr[r(\bv{x}_{t+1}, \bv{u}_{t+1}) + \gamma \int f(\bv{x}_{t+2}) \mathcal{J}_{t+2}d\bv{x}_{t+2}\biggl] d\bv{x}_{t+1},\\
    &= \phi(\hat{\bv{x}}_t,\bv{u}_t) + \gamma\int \hat{f}_{\theta_{b}}(\hat{\bv{x}}_{t+1})\phi(\hat{\bv{x}}_{t+1}, \bv{u}_{t+1})d\hat{\bv{x}}_{t+1}\notag\\
    &+ \gamma^2 \int \hat{f}_{\theta_{b}}(\hat{\bv{x}}_{t+1})\int \hat{f}_{\theta_{b}}(\hat{\bv{x}}_{t+2}) \phi(\hat{\bv{x}}_{t+2},\bv{u}_{t+2})d\hat{\bv{x}}_{t+2}d\hat{\bv{x}}_{t+1} +\notag\\
    & \dots + \gamma^{T-1} \int \hat{f}_{\theta_{b}}(\hat{\bv{x}}_{t+1})\dots\notag\\
    &\int \hat{f}_{\theta_{b}}(\hat{\bv{x}}_{t+T-1}) \nabla_{\hat{f}_{\theta_{b}}}\mathcal{J}_{t+T-1}d\hat{\bv{x}}_{t+T-1:t+1}.
\end{align}

Note that,
$\nabla_{\hat{f}_{\theta_{b}}}\mathcal{J}_{t+T-1}=\nabla_{\hat{f}_{\theta_{b}}}r(\hat{\bv{x}}_{t+T-1},\bv{u}_{t+T-1})=0.$ Then, by substituting $\phi$ in \eqref{eqn:gradientVF} we obtain:
\begin{equation}\label{eqn:gradientVF2}
\begin{split}
&\nabla_{\hat{f}_{\theta_{b}}}\mathcal{J}_{t} = \gamma \int \nabla_{\hat{f}_{\theta_{b}}} \hat{f}_{\theta_{b}}(\hat{\bv{x}}_{t+1})\mathcal{J}_{t+1}d\hat{\bv{x}}_{t+1}\\
    & + \gamma^2 \int \hat{f}_{\theta_{b}}(\hat{\bv{x}}_{t+1})\int \nabla_{\hat{f}_{\theta_{b}}} (\hat{f}_{\theta_{b}}(\hat{\bv{x}}_{t+2}))\mathcal{J}_{t+2}d\hat{\bv{x}}_{t+2:t+1}\\
    & +\dots + \gamma^{T-1} \int \hat{f}_{\theta_{b}}(\hat{\bv{x}}_{t+1})\dots\\
    &\int \nabla_{\hat{f}_{\theta_{b}}}(\hat{f}_{\theta_{b}}(\hat{\bv{x}}_{t+T-1}))\mathcal{J}_{t+T-1}d\hat{\bv{x}}_{t+T-1:t+1}.
\end{split}
\end{equation}

Multiplying each term in \eqref{eqn:gradientVF2} by $\frac{\hat{f}_{\theta_{b}}(\hat{\bv{x}}_{t+i})}{\hat{f}_{\theta_{b}}(\hat{\bv{x}}_{t+i})}$, and observing that $\int \hat{f}_{\theta_{b}}(\hat{\bv{x}}_{t+1})\nabla_{\hat{f}_{\theta_{b}}} \ln(\hat{f}_{\theta_{b}}(\hat{\bv{x}}_{t+1}))\mathcal{J}_{t+1}d\hat{\bv{x}}_{t+1} = \mathbb{E}_{\hat{\bv{x}}_{t+1}}\biggr[ \nabla_{\hat{f}_{\theta_{b}}} \ln(\hat{f}_{\theta_{b}}(\hat{\bv{x}}_{t+1}))\mathcal{J}_{t+1}\biggl]$, we can write:

\begin{align}\label{eqn:gradientVF3}
    &\nabla_{\hat{f}_{\theta_{b}}}\mathcal{J}_{t}= \gamma \mathbb{E}_{\hat{\bv{x}}_{t+1}}\biggr[ \nabla_{\hat{f}_{\theta_{b}}} \ln(\hat{f}_{\theta_{b}})\mathcal{J}_{t+1} \biggl]\notag\\ 
    &+ \gamma^2 \mathbb{E}_{\hat{\bv{x}}_{t+1:t+2}\sim\hat{f}_{\theta_{b}}}\biggr[ \nabla_{\hat{f}_{\theta_{b}}} \ln(\hat{f}_{\theta_{b}}(\hat{\bv{x}}_{t+2}))\mathcal{J}_{t+2}\biggl]+\dots\notag\\
    &+ \gamma^{T-1} \mathbb{E}_{\hat{\bv{x}}_{t+1:t+T-1}\sim\hat{f}_{\theta_{b}}}\biggr[ \nabla_{\hat{f}_{\theta_{b}}} \ln(\hat{f}_{\theta_{b}}(\hat{\bv{x}}_{t+T-1}))\mathcal{J}_{t+T-1} \biggl]\notag\\
    &= \sum\limits_{k=1}^{T-1} \gamma^{k} \mathbb{E}_{\hat{\bv{x}}_{t+1:t+k}\sim\hat{f}_{\theta_{b}}}\biggr[ \nabla_{\hat{f}_{\theta_{b}}} \ln(\hat{f}_{\theta_{b}}(\hat{\bv{x}}_{t+k}))\mathcal{J}_{t+k} \biggl]\notag\\
    % &= \sum\limits_{k=1}^{T-1} \gamma^{k} \mathbb{E}_{\hat{\bv{x}}_{t+1:t+k}\sim\hat{f}_{\theta_{b}}}\biggr[ \nabla_{\hat{f}_{\theta_{b}}} \ln(\hat{f}_{\theta_{b}}(\hat{\bv{x}}_{t+k}|\hat{\bv{x}}_{t+k-1}, \bv{u}_{t+k-1}))\\
    &\sum\limits_{i=t+k}^{t+T-1} \mathbb{E}_{\hat{\bv{x}}_{i+1}\sim\hat{f}_{\theta_{b}}}\bigr[\gamma^{i-t-k} r(\hat{\bv{x}}_{i}, \bv{u}_{i})\bigl] \biggl].
\end{align}

By combining \eqref{eqn:Jimmy} and \eqref{eqn:gradientVF3} we obtain \eqref{eqn:thm_statement}. \end{proof}

% \begin{algorithm}
% \caption{Distributionally robust probabilistic ensembles with trajectory sampling (DRPETS)}
% \begin{algorithmic}[1]
% \State Initialize data $\mathcal{D}$ with a random controller for one rollout.
% \For{Iteration $k = 1$ to $K$}
%     \State Train a probabilistic ensemble dynamics model $\hat{f}_{\theta}$ given data-set $\mathcal{D}$.
%     \For{Time $t = 0$ to $T$}
%         \For{Actions sampled $\bv{u}_{t:t+T} \sim \mathbf{CEM}(\cdot), 1$ to $N$ samples} 
%         \State Propagate state particles $\bv{x}_t^p$ using trajectory sampling and $\hat{f}_{\theta_{b}}\mid\{\mathcal{D}, \bv{u}_{t:t+T}\}$.
%         \State Evaluate actions as per \eqref{eqn:thm_statement}
%         \State Update $\mathbf{CEM}(\cdot)$ distribution.
%         \State Execute first action $\bv{u}_t^*$ (only) from optimal actions $\bv{u}_{t:t+T}^*$.
%         \State Record outcome: $\mathcal{D} \gets \mathcal{D} \cup \{\bv{x}_t, \bv{u}_t^*, \bv{x}_{t+1}\}$.
%         \EndFor
%     \EndFor
% \EndFor
% \end{algorithmic}
% \end{algorithm}

To recap, we aim to solve \eqref{eqn:MPC_prob} despite the unknown transition model \( f \). We address this by formulating a DR version of \eqref{eqn:Chua_obj}, assuming \( f \) follows a distribution within the ambiguity set \( \Omega_{\epsilon}(\hat{\mu}) \). Given the WR-MDP, we define the DR optimization problem in \eqref{eqn:DR_obj}.  
Applying \( p \)-Wasserstein duality, we derive \eqref{eqn:thm_1_statement} as a tractable solution to Problem \ref{prob1}, comprising an empirical estimate over the ensemble performance index and a regularizer. Leveraging the B-NNs ensemble, we further simplify Theorem \ref{thm:Jimmy}, yielding \eqref{eqn:thm_statement}.

\section{Numerical results}

% \grtodo{Here we go back to DR-PETS...but there is no mention to it in the contributions.}
% \AAC{We evaluate our DR-PETS algorithm under epistemic uncertainty, arising from deploying the learned policy on environments that differ from the nominal system used during training.} To investigate this, we conduct experiments on the Pendulum and Cartpole swing-up tasks, comparing our approach with the PETS algorithm~\cite{chua2018}. We develop custom environments that allow for perturbations in system parameters, thereby explicitly challenging the policy with unseen dynamics. For each task, a 200-step horizon is considered, and an ensemble of \( B = 5 \) neural networks is employed to learn the transition probability functions, with \( Q = 10 \) propagated particles. The B-NN ensemble is trained on the nominal environment for 100 episodes, after which we evaluate the robustness of the obtained DR-PETS policy over a range of perturbed environments. The code for the DR-PETS implementation with the custom environment is available at \url{https://tinyurl.com/595c76t6}.
We evaluate DR-PETS under epistemic uncertainty by testing its learned policy in perturbed environments distinct from the nominal training setup. Experiments on Pendulum and Cartpole swing-up tasks compare DR-PETS with PETS~\cite{chua2018}, using custom environments with parameter perturbations to challenge policy robustness. Each task runs for 200 steps with an ensemble of \( B = 5 \) neural networks modeling transition probabilities and \( Q = 10 \) propagated particles. The B-NN ensemble is trained on the nominal environment for 100 episodes, after which DR-PETS' robustness is assessed across perturbed settings. The implementation and custom environments are available at \url{https://tinyurl.com/595c76t6}.

% \grtodo{Credo dovremmo essere piu' espliciti su come definiamo le incertezze. La cosa interessante e' che facciamo deployment su un sistema differente da quello usato durante il training. La direi proprio cosi'.}
% \subsection{Example: Pendulum task}

% \grtodo{In questo paragrafo dettaglierei quale e' l'incertezza, legandolo al paragrafo precedente (deployment).}
\subsection{Example: Pendulum task}
We start with a nominal pendulum of mass 1 kg and length 0.5 m. Using the algorithm and code from \cite{chua2018}, we train an ensemble of 5 NNs to estimate the transition probability function and solve the MPC planning problem via CEM. DR-PETS differs from PETS by optimizing a distributionally robust MPC objective \eqref{eqn:thm_statement}, whereas PETS uses \eqref{eqn:Chua_obj}. To compare performance, we simulate pendulums with masses varying from 0.5 kg to 1.5 kg i.e. uncertainty set and evaluate total rewards across the set of masses.
% First, we begin with a nominal pendulum having a weight ($w$) of 1 kg and a length ($l$) of 0.5 m. Subsequently, we employ the algorithm and code provided by \cite{chua2018} to train an ensemble of 5 NNs that estimates the transition probability function. We solve the MPC planning problem using this NN ensemble in conjunction with the CEM. What distinguishes DR-PETS from PETS is the distributionally robust MPC objective. PETS optimizes actions ($u$) based on the objective function defined in \eqref{eqn:Chua_obj}, while DR-PETS tackles the objective outlined in \eqref{eqn:thm_statement}, namely  a DR objective function. To assess the performance of DR-PETS relative to PETS, we conduct simulations involving pendulums with varying masses, ranging from $0.5$ kg to $1.5$ kg i.e. the uncertainty set, and then assess the total rewards achieved for each mass value.
% \begin{figure}[h!]
%     \centering
%     \includegraphics[width=1\textwidth]{Figures/Pendulum_reward_DRvsNom_2.png}
%     \caption{Total episodic reward obtained by the PETS (in blue) and DR-PETS (in red) for perturbation of pendulum mass. Shaded region denotes one standard deviation.  }
%     \label{fig:pendulum_plot}
% \end{figure}

\begin{figure}[h!]
    \centering
    \includegraphics[width=0.5\textwidth]{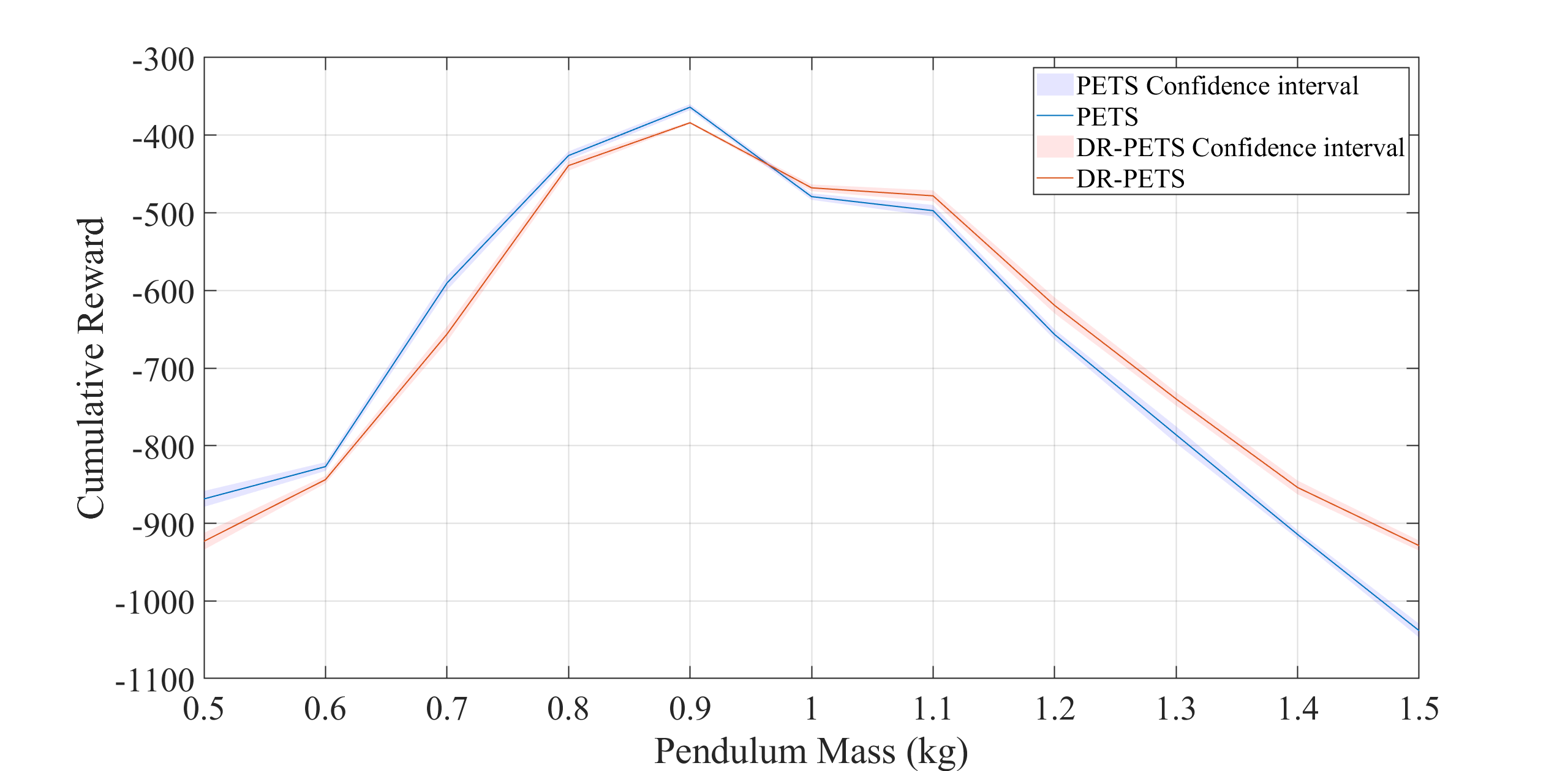}
    \caption{Total episodic reward obtained by the PETS (in blue) and DR-PETS (in red) for perturbation of pendulum mass. Shaded region denotes half of one standard error.}
    \label{fig:pendulum_plot}
\end{figure}

The plot in figure~\ref{fig:pendulum_plot} is obtained after 50 simulations for each mass perturbation. The performance difference between PETS and DR-PETS is under parameter perturbations is marginal, although the proposed algorithm exhibits more robustness to model parameter uncertainty. Indeed, when the mass value is higher than 1 kg, the DR-PETS reaches slight higher cumulative reward. %This benefit can become very important in more complex tasks.
%\grtodo{Questo paragrafo mi sembra contro producente. Dobbiamo commentare la figura ma non si capisce l'ultima frase...o siamo piu' precisi oppure rimuoviamola...}
\subsection{Example: Cartpole task}
\begin{figure}[h!]
    \centering
    \includegraphics[width=0.5\textwidth]{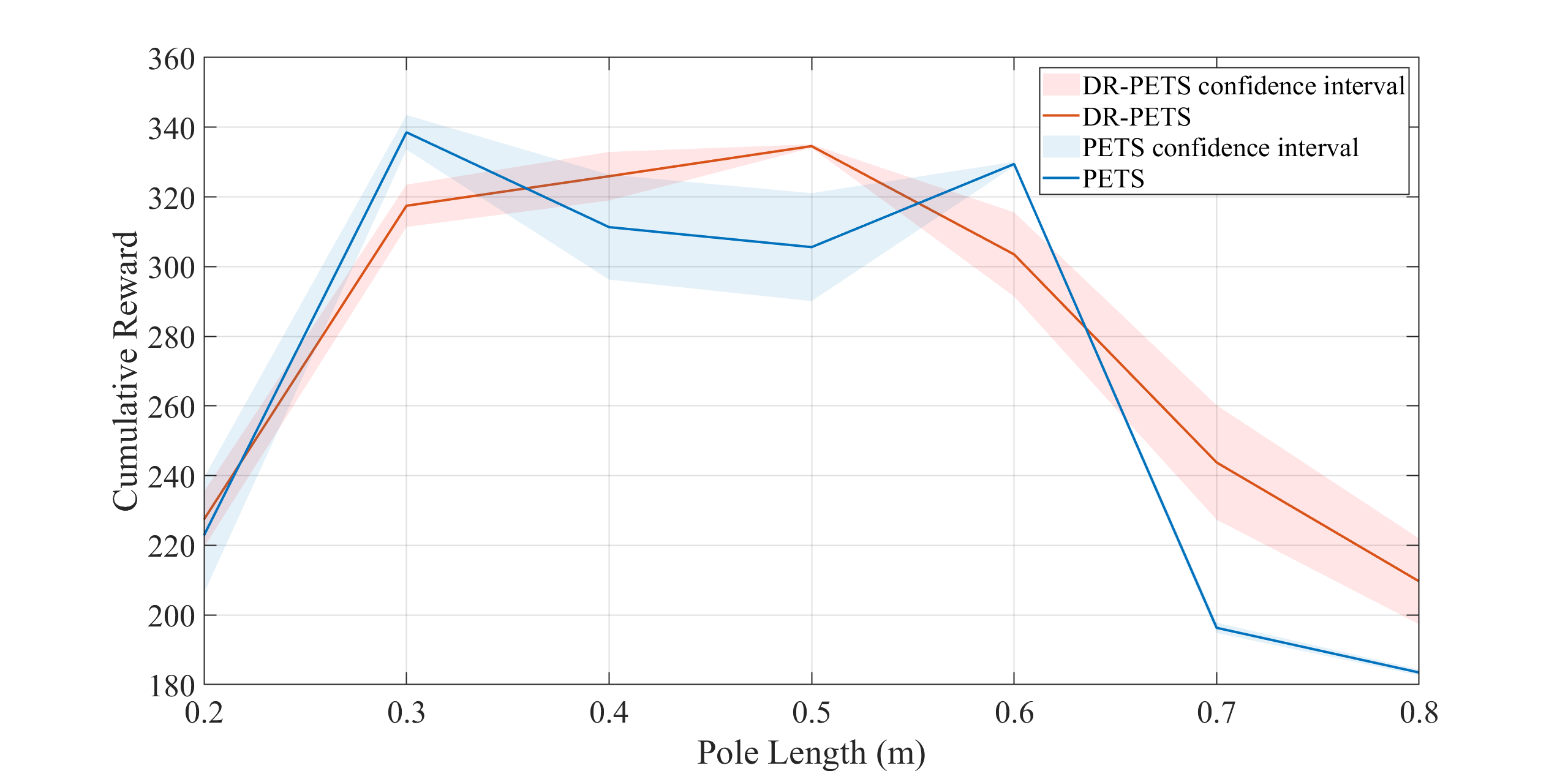}
    \caption{Total episodic reward obtained by the PETS (in blue) and DR-PETS (in red) for perturbation of pole length. Shaded region denotes half of one standard error.}
    \label{fig:cartpole_plot}
\end{figure}
We take a nominal cartpole system with the pole length ($l$) and mass ($m$) being 0.5 m and 1 kg respectively; we employ the algorithm and code provided by \cite{chua2018} to train an ensemble of 5 NNs that estimate the system dynamics. 
% We solve the MPC planning problem over a horizon ($T$) using this NN ensemble in conjunction with the cross-entropy optimizer. What distinguishes DR-PETS from PETS is the distributionally robust MPC objective. PETS optimizes actions ($u$) based on the objective function defined in \eqref{eqn:Chua_obj}, while DR-PETS tackles the objective outlined in \eqref{eqn:thm_statement}, which represents a distributionally robust objective function. 
To assess the performance of DR-PETS relative to PETS, we conduct simulations involving cartpole with varying pole lengths, ranging from 0.2 m to 0.8 m, and then assess the total rewards achieved for each length value.

Figure~\ref{fig:cartpole_plot} presents cumulative reward comparisons of DR-PETS and PETS over 50 simulations against length perturbation. Beyond a 0.6 m pole, both algorithms show a sharp performance decline, but DR-PETS degrades more gradually, maintaining higher cumulative rewards. Additionally, PETS exhibits a wider confidence interval, suggesting greater performance variability.
% The plot in figure~\ref{fig:cartpole_plot} is obtained after 50 simulations for each length perturbation. The plot shows comparison of DR-PETS with PETS for cumulative reward. Beyond the 0.6 m pole length, both algorithms experience a sharp decline in performance as the pole length increases. However, DR-PETS appears to have a slower rate of decline compared to PETS, resulting in DR-PETS having a higher cumulative reward than DR-PETS at pole lengths beyond 0.6 m. Also, the confidence interval for PETS is consistently wider than for DR-PETS, which might indicate that has a higher variability in its performance.
\section*{Acknowledgments}
This research would have been impossible without the foundational contributions and invaluable guidance of Yannis~Paschalidis and Jimmy~Queeney.  
\appendix
\subsection{Special cases: $p=\infty$ and $p=1$}
When $p=\infty$, Wasserstein duality results in 

\begin{equation}
\begin{aligned}
     &\frac{1}{B}\sum\limits_{b=1}^B\inf_{\tilde{f}_b}
    \bigr[\mathcal{J}_{t}(\tilde{f}_{b}) \mid d(\tilde{f}_b, \hat{f}_{\theta_{b}})\leq\epsilon\bigl]
    \\  &=  \frac{1}{B}\sum\limits_{b=1}^B\inf_{\norm{v_{b}}\leq\epsilon}
    \bigr[\mathcal{J}_{t}(\hat{f}_{\theta_{b}}+v_{b})\bigl]
\end{aligned}
\end{equation}

By considering a first order approximation of $\mathcal{J}(\cdot)$ with respect to the inputs $\tilde{f}$, we have

\begin{equation}
\begin{aligned}
    &\frac{1}{B}\sum\limits_{b=1}^B\inf_{\norm{v_{b}}\leq\epsilon}
    \bigr[\mathcal{J}_{t}(\hat{f}_{\theta_{b}}+v_{b})\bigl]
    % &=\frac{1}{B}\sum\limits_{b=1}^B\inf_{\norm{v_{b}}\leq\epsilon}\biggr[\mathcal{J}_{t}(\hat{f}_{\theta_{b}})+\left<\nabla_{\hat{f}_{\theta_{b}}}\mathcal{J}_{t}(\hat{f}_{\theta_{b}}),v_{b}\right>\biggl]\\
    % &=\frac{1}{B}\sum\limits_{b=1}^B\mathcal{J}_{t}(\hat{f}_{\theta_{b}})+\frac{1}{B}\sum\limits_{b=1}^B\inf_{\norm{v_{b}}\leq\epsilon}\biggr[\left<\nabla_{\hat{f}_{\theta_{b}}}\mathcal{J}_{t}(\hat{f}_{\theta_{b}}),v_{b}\right>\biggl]\\
    \\&=\frac{1}{B}\sum\limits_{b=1}^B\mathcal{J}_{t}(\hat{f}_{\theta_{b}})-\frac{\epsilon}{B}\sum\limits_{b=1}^B\norm{\nabla_{\hat{f}_{\theta_{b}}}\mathcal{J}_{t}(\hat{f}_{\theta_{b}})}_{*}.
    \end{aligned}
\end{equation}
As a result, the closed-form solution and adversarial perturbations follow the same form as in
Theorem \ref{thm:Jimmy}. Also note that $p = \infty$ results in the special case, where all inputs are perturbed using the same magnitude $\epsilon$.

%\subsection{Special case: $p=1$}
When $p=1$, all previous derivations hold through \eqref{eqn:delta_opt},

\begin{equation}\label{eqn:delta_opt_spcl}
    \begin{split}
    &\inf\limits_{\delta_{b}\geq0}\left\{\lambda\delta_{b}-\delta_{b}\norm{\nabla_{\hat{f}_{\theta_{b}}}\mathcal{J}_{t}(\hat{f}_{\theta_{b}})}_{*}\right\}\\
    &=\inf\limits_{\delta_{b}\geq0}\left\{\delta_{b}\left(\lambda-\norm{\nabla_{\hat{f}_{\theta_{b}}}\mathcal{J}_{t}(\hat{f}_{\theta_{b}})}_{*}\right)\right\},
    \end{split}
\end{equation}

We can see that in \eqref{eqn:delta_opt_spcl} equals zero when $\lambda\geq\norm{\nabla_{\hat{f}_{\theta_{b}}}\mathcal{J}_{t}(\hat{f}_{\theta_{b}})}_{*}$ and $-\infty$ otherwise. Then while considering the outer maximization problem
\begin{equation*}
    \sup_{\lambda \geq 0}\biggr\{\frac{1}{B}\sum\limits_{b=1}^B\inf_{v_{b}}\left[\left<\nabla_{\hat{f}_{\theta_{b}}}\mathcal{J}_{t}(\hat{f}_{\theta_{b}}),v_{b}\right> + \lambda\norm{v_{b}}\right] - \lambda \epsilon\biggl\},
\end{equation*}
we must have $\lambda^{*}=\max_{b=1,\dots,B}\norm{\nabla_{\hat{f}_{\theta_{b}}}\mathcal{J}_{t}(\hat{f}_{\theta_{b}})}_{*}$, which results in the closed-form solution for the Wasserstein distributionally robust objective to become

\begin{equation}\label{eqn:p=1 objective}
    \frac{1}{B}\sum\limits_{b=1}^B\mathcal{J}_{t}(\hat{f}_{\theta_{b}})-\epsilon\max_{b=1,\dots,B}\norm{\nabla_{\hat{f}_{\theta_{b}}}\mathcal{J}_{t}(\hat{f}_{\theta_{b}})}_{*}
\end{equation}

The form of \eqref{eqn:p=1 objective} resembles Theorem \ref{thm:Jimmy}. However, there is no adversarial interpretation when $p = 1$,
as the magnitude of each adversarial perturbation is chosen to be zero in \eqref{eqn:delta_opt_spcl}.
% \begin{figure}[h!]
%     \centering
%     \includegraphics[width=0.5\textwidth]{Figures/cartpole_plot.png}
%     \caption{Total episodic reward obtained by the PETS (in blue) and DR-PETS (in red) for perturbation of pole length. Shaded region denotes half of one standard error.}
%     \label{fig:cartpole_plot}
% \end{figure}

\bibliographystyle{ieeetr}
\bibliography{root_sub}

\end{document}